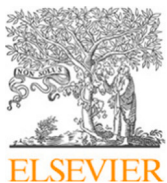
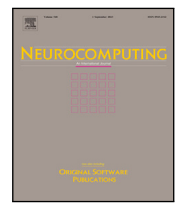

# KeyCLD: Learning constrained Lagrangian dynamics in keypoint coordinates from images

Rembert Daems [a,b,c,*], Jeroen Taets [a,c], Francis wyffels [b], Guillaume Crevecoeur [a,c]

[a] *D2LAB – Ghent University, Tech Lane Ghent Science Park 131, Zwijnaarde, 9052, Belgium*
[b] *IDLab-AIRO – Ghent University – imec, Tech Lane Ghent Science Park 126, Zwijnaarde, 9052, Belgium*
[c] *MIRO core lab – Flanders Make@UGent, Tech Lane Ghent Science Park 131, Zwijnaarde, 9052, Belgium*

## ARTICLE INFO

Communicated by J. Chen

*Keywords:*
Lagrangian
Dynamics
Video
Images
Unsupervised learning
Energy shaping control

## ABSTRACT

We present KeyCLD, a framework to learn Lagrangian dynamics from images. Learned keypoints represent semantic landmarks in images and can directly represent state dynamics. We show that interpreting this state as Cartesian coordinates, coupled with explicit holonomic constraints, allows expressing the dynamics with a constrained Lagrangian. KeyCLD is trained unsupervised end-to-end on sequences of images. Our method explicitly models the mass matrix, potential energy and the input matrix, thus allowing energy based control. We demonstrate learning of Lagrangian dynamics from images on the `dm_control` pendulum, cartpole and acrobot environments. KeyCLD can be learned on these systems, whether they are unactuated, underactuated or fully actuated. Trained models are able to produce long-term video predictions, showing that the dynamics are accurately learned. We compare with Lag-VAE, Lag-caVAE and HGN, and investigate the benefit of the Lagrangian prior and the constraint function. KeyCLD achieves the highest valid prediction time on all benchmarks. Additionally, a very straightforward energy shaping controller is successfully applied on the fully actuated systems.

## 1. Introduction and related work

Learning dynamical models from data is a crucial aspect while striving towards intelligent agents interacting with the physical world. Understanding the dynamics and being able to predict future states is paramount for controlling autonomous systems or robots interacting with their environment. For many dynamical systems, the equations of motion can be derived from scalar functions such as the Lagrangian or Hamiltonian. This strong physics prior enables more data-efficient learning and holds energy conserving properties. Greydanus et al. [1] introduced Hamiltonian neural networks. By using Hamiltonian mechanics as inductive bias, the model respects exact energy conservation laws. Lutter et al. [2,3] pioneered the use of Lagrangian mechanics as physics priors for learning dynamical models from data. Cranmer et al. [4] expanded this idea to a more general setting. By modelling the Lagrangian itself with a neural network instead of explicitly modelling mechanical kinetic energy, they can model physical systems beyond classical mechanics. Finzi et al. [5] introduced learning of Lagrangian or Hamiltonian dynamics in Cartesian coordinates, with explicit constraints. This enables more data efficient models, at the cost of providing extra knowledge about the system in the form of a constraint function.

Zhong et al. [6] included external input forces and energy dissipation, and introduced energy-based control by leveraging the learned energy models.

### 1.1. Learning Lagrangian dynamics from images

It is often not possible to observe the full state of a system directly. Cameras provide a rich information source, containing the full state when properly positioned. However, the difficulty lies in interpreting the images and extracting the underlying state. As was recently argued by Lutter and Peters [7], learning Lagrangian or Hamiltonian dynamics from realistic images remains an open challenge. The majority of related work [1,8–11] use a variational autoencoder (VAE) framework to represent the state in a latent space embedding. The dynamics model is expressed in this latent space. Zhong and Leonard [12] use interpretable coordinates, however their state estimator module needs full knowledge of the kinematic chain, and the images are segmented per object. Table 1 provides an overview of closely related work in literature.

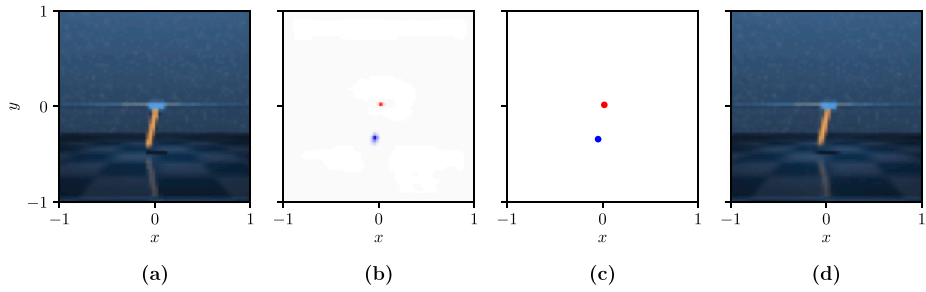

**Fig. 1.** KeyCLD learns Lagrangian dynamics from images. **(a)** An observation of a dynamical system is processed by a keypoint estimator model. **(b)** The model represents the positions of the keypoints with a set of spatial probability heatmaps. **(c)** Cartesian coordinates are extracted using spatial softmax and used as state representations to learn Lagrangian dynamics. **(d)** The information in the keypoint coordinates bottleneck suffices for a learned renderer model to reconstruct the original observation, including background, reflections and shadows. The keypoint estimator model, Lagrangian dynamics models and renderer model are jointly learned unsupervised on sequences of images.

**Table 1**
An overview of closely related Lagrangian or Hamiltonian models. Lag-caVAE [12] is capable of modelling external forces and learning from images, but individual moving bodies need to be segmented in the images, on a black background. It additionally needs full knowledge of the kinematic chain, which is more prior information than the constraint function for our method (see Section 2.3). HGN [8] needs no prior knowledge of the kinematic chain, but is unable to model external forces. CHNN [5] expresses Lagrangian or Hamiltonian dynamics in Cartesian coordinates, but cannot be learned from images. Our method, KeyCLD, is capable of learning Lagrangian dynamics with external forces, from unsegmented images with shadows, reflections and backgrounds.

|  | HGN | Lag-caVAE | CHNN | KeyCLD |
| --- | --- | --- | --- | --- |
| External forces (control) |  | ✓ |  | ✓ |
| Interpretable coordinates |  | ✓ | ✓ | ✓ |
| Cartesian coordinates |  |  | ✓ | ✓ |
| Learns from images | ✓ | ✓ |  | ✓ |
| Learns from unsegmented images | ✓ |  |  | ✓ |
| Needs kinematic chain prior |  | ✓ |  |  |
| Needs constraint prior |  |  | ✓ | ✓ |

### 1.2. Keypoints

Instead of using abstract latent embeddings, our method leverages fully convolutional keypoint estimator models to observe the state from images. Objects can be represented with one or more keypoints, fully capturing the position and orientation. Because the model is fully convolutional, it is also translation equivariant, hence exhibiting higher data efficiency. Zhou et al. [13] used keypoints for object detection, with great success. Keypoint detectors are commonly used for human pose estimation [14]. More closely related to this work, keypoints can be learned for control and robotic manipulation [15,16]. Minderer et al. [17] learn unsupervised keypoints from videos to represent objects and dynamics. Jaques et al. [18] leverage keypoints for system identification and dynamic modelling. Jakab et al. [19] learn a keypoint representation unsupervised by using it as an information bottleneck for reconstructing images. The keypoints represent semantic landmarks in the images and generalize well to unseen data. It is the main inspiration for the use of keypoints in our work.

### 1.3. Contributions

Concretely, our work makes the following contributions. **(1)** We introduce KeyCLD, a framework to learn constrained Lagrangian dynamics from images. We are the first to use learned keypoint representations from images to learn Lagrangian dynamics. We show that keypoint representations derived from images can directly be used as positional state vector for constrained Lagrangian dynamics, expressed in Cartesian coordinates. **(2)** We show how to control constrained Lagrangian dynamics in Cartesian coordinates with energy shaping, where the state is estimated from images. **(3)** KeyCLD is empirically validated on the pendulum, cartpole and acrobot systems from `dm_control` [20]. We show that KeyCLD can be learned on these systems, whether they are unactuated, underactuated or fully actuated. We compare quantitatively with Lag-caVAE, Lag-VAE [12] and HGN [8], and investigate the benefit of the Lagrangian prior and the constraint function. KeyCLD achieves the highest valid prediction time on all benchmarks (see Table 2).

## 2. Constrained Lagrangian dynamics

### 2.1. Lagrangian dynamics

For a dynamical system with $m$ degrees of freedom, a set of independent generalized coordinates $\mathbf{q} \in \mathbb{R}^m$ represents all possible kinematic configurations of the system. The time derivatives $\dot{\mathbf{q}} \in \mathbb{R}^m$ are the velocities of the system. If the system is fully deterministic, its dynamics can be described by the equations of motion, a set of second order ordinary differential equations (ODE):

$$\ddot{\mathbf{q}} = \mathbf{f}(\mathbf{q}(t), \dot{\mathbf{q}}(t), t, \mathbf{u}(t)) \qquad (1)$$

where $\mathbf{u}(t)$ are the external forces acting on the system. From a known initial value $(\mathbf{q}, \dot{\mathbf{q}})$, we can integrate $\mathbf{f}$ through time to predict future states of the system. It is possible to model $\mathbf{f}$ with a neural network, and train the parameters with backpropagation through an ODE solver [21].

However, by expressing the dynamics with a Lagrangian we introduce a strong physics prior [3]:

$$\mathcal{L}(\mathbf{q}, \dot{\mathbf{q}}) = T(\mathbf{q}, \dot{\mathbf{q}}) - V(\mathbf{q}) \qquad (2)$$

where $T$ is the kinetic energy and $V$ is the potential energy of the system. For any mechanical system the kinetic energy is defined as:

$$T(\mathbf{q}, \dot{\mathbf{q}}) = \frac{1}{2}\dot{\mathbf{q}}^\top \mathbf{M}(\mathbf{q})\dot{\mathbf{q}} \qquad (3)$$

where $\mathbf{M}(\mathbf{q}) \in \mathbb{R}^{m \times m}$ is the positive semi-definite mass matrix. Ensuring that $\mathbf{M}(\mathbf{q})$ is positive semi-definite can be done by expressing $\mathbf{M}(\mathbf{q}) = \mathbf{L}(\mathbf{q})\mathbf{L}(\mathbf{q})^\top$, where $\mathbf{L}(\mathbf{q})$ is a lower triangular matrix. It is now possible to describe the dynamics with two neural networks, one for the mass matrix and one for the potential energy. Since both are only in function of $\mathbf{q}$ and not $\dot{\mathbf{q}}$, and expressing the mass matrix and potential energy is more straightforward than expressing the equations of motion, it is generally much more simple to learn dynamics with this framework. In other words, adding more physics priors in the form of Lagrangian mechanics, makes learning the dynamics more robust and data-efficient [2–4,7].

The Euler–Lagrange equations (4) allow transforming the Lagrangian into the equations of motion by solving for $\ddot{\mathbf{q}}$:

$$\frac{d}{dt}\nabla_{\dot{\mathbf{q}}}\mathcal{L} - \nabla_{\mathbf{q}}\mathcal{L} = \nabla_{\mathbf{q}}W \qquad (4)$$

$$\nabla_{\mathbf{q}}W = \mathbf{g}(\mathbf{q})\mathbf{u} \qquad (5)$$

where $W$ is the external work done on the system, e.g. forces applied for control. The input matrix $\mathbf{g} \in \mathbb{R}^{m \times l}$ allows introducing external

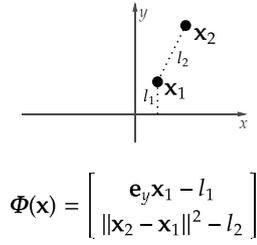

$$\Phi(\mathbf{x}) = \begin{bmatrix} \mathbf{e}_y \mathbf{x}_1 - l_1 \\ \|\mathbf{x}_2 - \mathbf{x}_1\|^2 - l_2 \end{bmatrix}$$

**Fig. 2.** Example of a constraint function $\Phi(\mathbf{x})$ to express the cartpole system with Cartesian coordinates. The cartpole system has 2 degrees of freedom, but is expressed in $\mathbf{x} \in \mathbb{R}^4$. Valid configurations of the system in $\mathbb{R}^4$ are constrained on a manifold defined by $\mathbf{0} = \Phi(\mathbf{x})$. The first constraint only allows horizontal movement of $\mathbf{x}_1$, and the second constraint enforces a constant distance between $\mathbf{x}_1$ and $\mathbf{x}_2$. Although unknown $l_1$ and $l_2$ constants are present in $\Phi(\mathbf{x})$, their values are irrelevant, since only the Jacobian of $\Phi(\mathbf{x})$ is used in our framework (see Eq. (13)). See Appendix D for more examples of constraint functions.

forces $\mathbf{u} \in \mathbb{R}^l$ for modelling any control-affine system. If the external forces and torques are aligned with the degrees of freedom $\mathbf{q}$, $\mathbf{g}$ can be a diagonal matrix or even an identity matrix. More generally, if no prior knowledge is present about the relationship between $\mathbf{u}$ and the generalized coordinates $\mathbf{q}$, $\mathbf{g}(\mathbf{q}) : \mathbb{R}^m \to \mathbb{R}^{m \times l}$ is a function of $\mathbf{q}$ and can be modelled with a third neural network [6]. If the system is fully actuated, the number of actuators equals the number of degrees of freedom ($l = m$), if it is underactuated not all degrees of freedom are actuated ($l < m$).

### 2.2. Cartesian coordinates

Finzi et al. [5] showed that expressing Lagrangian mechanics in Cartesian coordinates $\mathbf{x} \in \mathbb{R}^k$ instead of independent generalized coordinates $\mathbf{q} \in \mathbb{R}^m$ has several advantages:

$$\mathcal{L}(\mathbf{x}, \dot{\mathbf{x}}) = \frac{1}{2} \dot{\mathbf{x}}^\top \mathbf{M} \dot{\mathbf{x}} - V(\mathbf{x}) \quad (6)$$

The mass matrix $\mathbf{M}$ no longer changes in function of the state, and is thus static. This means that a neural network is no longer required to model the mass matrix, simply the values in the matrix itself are optimized. The potential energy $V(\mathbf{x})$ is now in function of $\mathbf{x}$. Expressing the potential energy in Cartesian coordinates can often be simpler than in generalized coordinates. E.g. for gravity, this is simply a linear function. Also the input matrix $\mathbf{g}(\mathbf{x})$ is now in function of $\mathbf{x}$, and the shape is $k \times l$. Multiplying the input matrix with the input vector $\mathbf{g}(\mathbf{x})\mathbf{u}$ results in force vectors acting in Cartesian space. The input matrix is modelled with a fully connected neural network, of which the output vector is reshaped in the correct shape.

Because we are now expressing the system in Cartesian coordinates we additionally need a set of $n$ holonomic constraint functions $\Phi(\mathbf{x}) : \mathbb{R}^k \to \mathbb{R}^n$. These guarantee a valid configuration of the system, and a correct number of degrees of freedom: $m = k - n$ (see Fig. 2). The constrained Euler–Lagrange equations are expressed with a vector $\lambda(t) \in \mathbb{R}^n$ containing Lagrange multipliers for the constraints [5,22]:

$$\frac{d}{dt} \nabla_{\dot{\mathbf{x}}} \mathcal{L}(\mathbf{x}, \dot{\mathbf{x}}) - \nabla_{\mathbf{x}} \mathcal{L}(\mathbf{x}, \dot{\mathbf{x}}) = \mathbf{g}(\mathbf{x})\mathbf{u}(t) + D\Phi(\mathbf{x})^\top \lambda(t) \quad (7)$$

with $D$ being the Jacobian operator. Because the mass matrix is static,[1] this is simplified to:

$$\mathbf{M}\ddot{\mathbf{x}} + \nabla_{\mathbf{x}} V(\mathbf{x}) = \mathbf{g}(\mathbf{x})\mathbf{u}(t) + D\Phi(\mathbf{x})^\top \lambda(t) \quad (8)$$

$$\ddot{\mathbf{x}} = \mathbf{M}^{-1}\mathbf{f} + \mathbf{M}^{-1} D\Phi(\mathbf{x})^\top \lambda(t), \quad \mathbf{f} = -\nabla_{\mathbf{x}} V(\mathbf{x}) + \mathbf{g}(\mathbf{x})\mathbf{u}(t) \quad (9)$$

---
[1] In other words, the centrifugal and Coriolis forces are zero because $\dot{\mathbf{M}} = \mathbf{0}$ and $\nabla_{\mathbf{x}} \mathbf{M} = \mathbf{0}$.

Calculating twice the time derivative of the constraint conditions yields:

$$\begin{aligned} \mathbf{0} &\equiv \Phi(\mathbf{x}) \\ \mathbf{0} &= \dot{\Phi}(\mathbf{x}) \\ \mathbf{0} &= D\Phi(\mathbf{x})\dot{\mathbf{x}} \\ \mathbf{0} &= D\dot{\Phi}(\mathbf{x})\dot{\mathbf{x}} + D\Phi(\mathbf{x})\ddot{\mathbf{x}} \end{aligned} \quad (10)$$

The Lagrange multipliers $\lambda(t)$ are solved by substituting $\ddot{\mathbf{x}}$ from Eq. (9) in Eq. (10):

$$\begin{aligned} -D\dot{\Phi}(\mathbf{x})\dot{\mathbf{x}} &= D\Phi(\mathbf{x})\mathbf{M}^{-1}\mathbf{f} + D\Phi(\mathbf{x})\mathbf{M}^{-1} D\Phi(\mathbf{x})^\top \lambda(t) \\ \lambda(t) &= \left[ D\Phi(\mathbf{x})\mathbf{M}^{-1} D\Phi(\mathbf{x})^\top \right]^{-1} \left[ D\Phi(\mathbf{x})\mathbf{M}^{-1}\mathbf{f} + D\dot{\Phi}(\mathbf{x})\dot{\mathbf{x}} \right] \end{aligned} \quad (11)$$

We use the chain rule a second time to get rid of the time derivative of $D\Phi(\mathbf{x})$:

$$D\dot{\Phi}(\mathbf{x})\dot{\mathbf{x}} = \langle D^2 \Phi, \dot{\mathbf{x}} \rangle \dot{\mathbf{x}} \quad (12)$$

Substituting $\lambda(t)$ in (9) we finally arrive at:

$$\begin{aligned} \mathbf{f} &= -\nabla_{\mathbf{x}} V + \mathbf{g}\mathbf{u} \\ \ddot{\mathbf{x}} &= \mathbf{M}^{-1}\mathbf{f} - \mathbf{M}^{-1} D\Phi^\top \left[ D\Phi \mathbf{M}^{-1} D\Phi^\top \right]^{-1} \left[ D\Phi \mathbf{M}^{-1} \mathbf{f} + \langle D^2 \Phi, \dot{\mathbf{x}} \rangle \dot{\mathbf{x}} \right] \end{aligned} \quad (13)$$

Since time derivatives of functions modelled with neural networks are no longer present, Eq. (13) can be implemented in an autograd library which handles the calculation of gradients and Jacobians automatically. See Appendix A for details and the implementation of Eq. (13) in JAX [23].

Note that in Eq. (13) only the Jacobian of $\Phi(\mathbf{x})$ is present. This means that there is no need to learn explicit constants in $\Phi(\mathbf{x})$, such as lengths or distances between points. Rather that constant distances and lengths through time are enforced by $D\Phi(\mathbf{x})\dot{\mathbf{x}} = \mathbf{0}$. We use this property to our advantage since this simplifies the learning process. See Fig. 2 for an example of the cartpole system expressed in Cartesian coordinates and a constraint function.

### 2.3. Constraints as prior knowledge

The given constraint function $\Phi(\mathbf{x})$ adds extra prior information to our model. Alternatively, we could use a mapping function $\mathbf{x} = \mathbf{F}(\mathbf{q})$. This leads directly to an expression of the Lagrangian in Cartesian coordinates using $\dot{\mathbf{x}} = D\mathbf{F}(\mathbf{q})\dot{\mathbf{q}}$:

$$\mathcal{L}(\mathbf{q}, \dot{\mathbf{q}}) = \frac{1}{2} \dot{\mathbf{q}}^\top D\mathbf{F}(\mathbf{q})^\top \mathbf{M} D\mathbf{F}(\mathbf{q}) \dot{\mathbf{q}} - V(\mathbf{F}(\mathbf{q})) \quad (14)$$

from which the equations of motion can be derived using the Euler–Lagrange equations, similar to Eq. (13). In terms of explicit knowledge about the system, the mapping $\mathbf{x} = \mathbf{F}(\mathbf{q})$ is equivalent to the kinematic chain as required for the method of Zhong and Leonard [12]. Using the constraint function is however more general. Some systems, such as systems with closed loop kinematics, cannot be expressed in generalized coordinates $\mathbf{q}$, and thus have no mapping function [24]. Furthermore, learning the constraint manifold $\mathbf{0} = \Phi(\mathbf{x})$ from data with geometric manifold learning algorithms [25,26] could be a future research direction. We therefore argue that adopting the constraint function $\Phi(\mathbf{x})$ is more general and requires less explicit knowledge injected in the model.

### 2.4. Relationship between Lagrangian and Hamiltonian

Both Lagrangian and Hamiltonian mechanics ultimately express the dynamics in terms of kinetic and potential energy. The Hamiltonian expresses the total energy of the system $H(\mathbf{q}, \mathbf{p}) = T(\mathbf{q}, \mathbf{p}) + V(\mathbf{q})$ [1,8]. It is expressed in the position and the generalized momenta $(\mathbf{q}, \mathbf{p})$, instead of generalized velocities. Using the Legendre transformation it is possible to transform $L$ into $H$ or back. We focus in our work on Lagrangian mechanics because it is more general [4] and observing the momenta $\mathbf{p}$ is impossible from images. See also Botev et al. [11] for a short discussion on the differences.

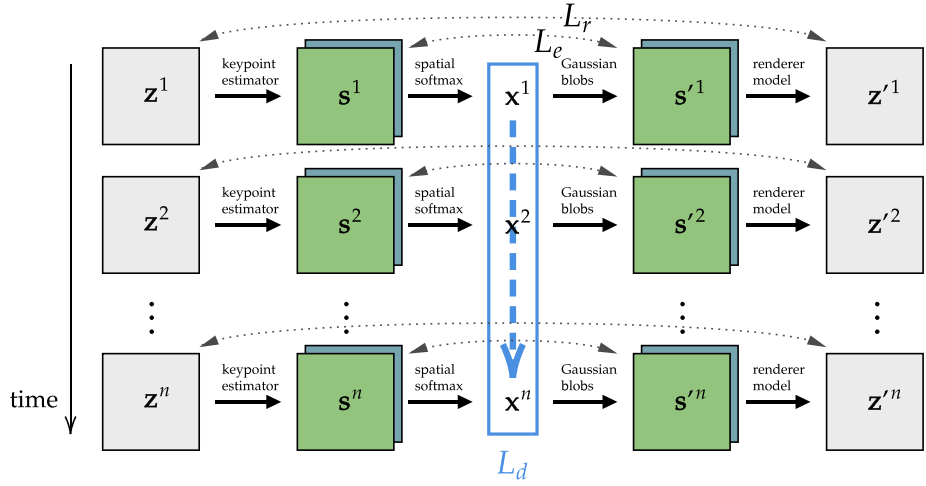

**Fig. 3.** Schematic overview of training KeyCLD. A sequence of $n$ images $\{z^i\}, i \in \{1, \ldots, n\}$, is processed by the keypoint estimator model, returning heatmaps $\{s^i\}$ representing spatial probabilities of the keypoints. $s^i$ consists of $m$ heatmaps $s_k^i$, one for every keypoint $x_k^i, k \in \{1, \ldots, m\}$. Spatial softmax is used to extract the Cartesian coordinates of the keypoints, and all keypoints are concatenated in the state vector $x^i$. $x^i$ is transformed back to a spatial representation $s'^i$ using Gaussian blobs. This prior is encouraged on the keypoint estimator model by a binary cross-entropy loss $L_e$ between $s^i$ and $s'^i$. The renderer model reconstructs images $z'^i$ based on $s'^i$, with reconstruction loss $L_r$. The dynamics loss $L_d$ is calculated on the sequence of state vectors $x^i$. Keypoint estimator model, renderer model and the dynamics models (mass matrix, potential energy and input matrix) are trained jointly with a weighted sum of the losses $L = L_r + L_e + \lambda L_d$.

## 3. Learning Lagrangian dynamics from images

### 3.1. Keypoints as state representations

We introduce the use of keypoints to learn Lagrangian dynamics from images. KeyCLD is trained unsupervised on sequences of $n$ images $\{z^i\}, i \in \{1, \ldots, n\}$ and a constant input vector $u$. See Fig. 3 for a schematic overview.

All images $z^i$ in the sequence are processed by the keypoint estimator model, returning each a set of heatmaps $s^i$ representing the spatial probabilities of keypoint positions. $s^i$ consists of $m$ heatmaps $s_k^i$, one for every keypoint $x_k^i, k \in \{1, \ldots, m\}$. The keypoint estimator model is a fully convolutional neural network, maintaining a spatial representation from input to output (see Fig. 4 for the detailed architecture). This contrasts with a model ending in fully connected layers regressing to the coordinates directly, where the spatial representation is lost [8,12]. Because a fully convolutional model is equivariant to translation, it can better generalize to unseen states that are translations of seen states. Another advantage is the possibility of augmenting $z$ with random transformations of the $D_4$ dihedral group to increase robustness and data efficiency. Because $s$ can be transformed back with the inverse transformation, this augmentation is confined to the keypoint estimator model and has no effect on the dynamics.

To distill keypoint coordinates from the heatmaps, we define a Cartesian coordinate system in the image (see for example Fig. 1). Based on this definition, every pixel $p$ corresponds to a point $x_p$ in the Cartesian space. The choice of the Cartesian coordinate system is arbitrary but is equal to the space of the dynamics $\ddot{x}(\dot{x}, x, t, u)$ and the constraint function $\Phi(x)$ (see Section 2). We use spatial softmax over all pixels $p \in \mathcal{P}$ to distill the coordinates of keypoint $x_k$ from its probability heatmap:

$$x_k = \frac{\sum_{p \in \mathcal{P}} x_p e^{s_k(p)}}{\sum_{p \in \mathcal{P}} e^{s_k(p)}} \quad (15)$$

Spatial softmax is differentiable, and the loss will backpropagate through the whole heatmap since $x_k$ depends on all the pixels. Cartesian coordinates $x_k$ of the different keypoints are concatenated in vector $x$ which serves as the state representation of the system. *This compelling connection between image keypoints and Cartesian coordinates forms the basis of this work.* The keypoint estimator model serves directly as state estimator to learn constrained Lagrangian dynamics from images.

Similar to Jakab et al. [19], $x$ acts as an information bottleneck, through which only the Cartesian coordinates of the keypoints flow to reconstruct the image with the renderer model. First, all $x_k$ are transformed back to spatial representations $s'_k$ using unnormalized Gaussian blobs, parameterized by a hyperparameter $\sigma$.

$$s'_k = \exp\left(-\frac{\|x_p - x_k\|^2}{2\sigma^2}\right) \quad (16)$$

A binary cross-entropy loss $L_e$ is formulated over $s$ and $s'$ to encourage this Gaussian prior. The renderer model can more easily interpret the state in this spatial representation, as it lies closer to its semantic meaning of keypoints as semantic landmarks in the reconstructed image. The renderer model learns a constant feature tensor (inspired by Nguyen-Phuoc et al. [27]), which provides it with positional information. Since the model itself is translation equivariant, it needs positional information to reconstruct background information or specific appearances that depend on the positions of objects. See Fig. 4 for the detailed architecture.

Finally, a reconstruction loss is formulated over the reconstructed images $z'^i$ and original images $z^i$:

$$L_r = \sum_{i=1}^{n} \|z'^i - z^i\|^2 \quad (17)$$

### 3.2. Dynamics loss function

The sequence $\{x^i\}$, corresponding to the sequence of given images $\{z^i\}$, and the constant input $u$ is used to calculate the dynamics loss. A fundamental aspect in learning dynamics from images is that velocities cannot be directly observed. A single image only captures the position of a system, and contains no information about its velocities.[2] Other work uses sequences of images as input to a model [8] or a specific velocity estimator model trained to estimate velocities from a sequence of positions [30]. Zhong and Leonard [12] demonstrate that for estimating velocities, finite differencing simply works better.

We use a central first order finite difference estimation, and project the estimated velocity on the constraints, so that the constraints are not

---

[2] Neglecting side-effects such as motion blur, which are not very useful for this purpose.

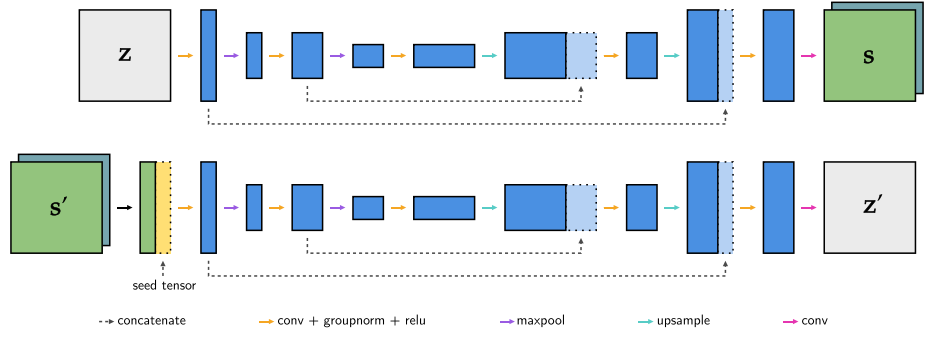

**Fig. 4.** Visualization of the keypoint estimator (top) and renderer (bottom) model architectures. The keypoint estimator model and renderer model have similar architectures, utilizing down- and upsampling and skip connections which help increasing the receptive field [28,29]. The renderer model learns a constant feature tensor that is concatenated with the input $\mathbf{s}'$. The feature tensor provides positional information since the fully-convolutional model is translation equivariant.

violated:

$$\dot{\mathbf{x}}^i = \left[I - D\Phi(\mathbf{x}^i)^+ D\Phi(\mathbf{x}^i)\right] \frac{\mathbf{x}^{i+1} - \mathbf{x}^{i-1}}{2h}, \quad i \in \{2, \ldots, n-1\} \quad (18)$$

where $(\cdot)^+$ signifies the Moore–Penrose pseudo-inverse and $h$ the timestep. We can now integrate future timesteps $\hat{\mathbf{x}}$ starting from initial values $(\mathbf{x}^i, \dot{\mathbf{x}}^i)$ using an ODE solver. The equations of motion (13) are solved starting from all initial values in parallel, for $\nu$ timesteps This maximizes the learning signal obtained to learn the dynamics and leads to overlapping sequences of length $\nu$:

$$\{\hat{\mathbf{x}}^{i+1}, \ldots, \hat{\mathbf{x}}^{i+\nu}\}, \quad i \in \{2, \ldots, n-\nu\} \quad (19)$$

Thus, $\hat{\mathbf{x}}^{i+j}$ is obtained by integrating $j$ timesteps forward in time, starting from initial value $\mathbf{x}^i$, which was derived by the keypoint estimator model. All $\{\hat{\mathbf{x}}^{i+j}\}$ in all sequences are compared with their corresponding keypoint states $\{\mathbf{x}^{i+j}\}$ in an $L_2$ loss:

$$L_d = \sum_{i=2}^{n-\nu} \sum_{j=1}^{\nu} \|\mathbf{x}^{i+j} - \hat{\mathbf{x}}^{i+j}\|^2 \quad (20)$$

### 3.3. Total loss

The total loss is the weighted sum of $L_r$, $L_e$ and $L_d$, with a weighing hyperparameter $\lambda$: $L = L_r + L_e + \lambda L_d$.

To conclude, the keypoint estimator model, renderer model and dynamics models (mass matrix, potential energy and input matrix) are jointly trained end-to-end on sequences of images $\{\mathbf{z}^i\}$ and constant inputs $\mathbf{u}$ with stochastic gradient descent.

### 3.4. Rigid bodies as rigid sets of point masses

By interpreting a set of keypoints as a set of point masses, we can represent any rigid body and its corresponding kinetic and potential energy. Additional constraints are added for the pairwise distances between keypoints representing a single rigid body [5]. For 3D systems, at least four keypoints are required to represent any rigid body [31]. We focus in our work on 2D systems in a plane parallel to the camera plane. 2D rigid bodies can be expressed with a set of 2 point masses, which can further be reduced depending on the constraints and connections between bodies (see Appendix B for more detail and proof). In our framework, the keypoint model is free to choose the relative placement of keypoints on the different moving parts of the dynamic system, enabling the choice of distinct landmarks that also express the state accurately, e.g. the endpoint of a beam.

The interpretation of rigid bodies as sets of point masses allows expressing the kinetic energy as the sum of the kinetic energies of the point masses. Corresponding to Eq. (3), the mass matrix for a 2D system is defined as a diagonal matrix with masses $m_k$ for every keypoint $\mathbf{x}_k$:

$$T(\dot{\mathbf{x}}) = \frac{1}{2}\dot{\mathbf{x}}^\top \mathbf{M}\dot{\mathbf{x}} = \frac{1}{2}\begin{bmatrix}\dot{\mathbf{x}}_1 & \cdots & \dot{\mathbf{x}}_n\end{bmatrix} \begin{bmatrix} m_1 & 0 & \cdots & 0 & 0 \\ 0 & m_1 & \cdots & 0 & 0 \\ \vdots & \vdots & \ddots & \vdots & \vdots \\ 0 & 0 & \cdots & m_n & 0 \\ 0 & 0 & \cdots & 0 & m_n \end{bmatrix} \begin{bmatrix}\dot{\mathbf{x}}_1 \\ \vdots \\ \dot{\mathbf{x}}_n\end{bmatrix} \quad (21)$$

To enforce positive values, the masses are parameterized by their square root and squared.

## 4. Experiments

We adapted the pendulum, cartpole and acrobot environments from `dm_control` [20,32] for our experiments. See Appendix D for details about the environments, their constraint functions and the data generation procedure. The exact same model architectures, hyperparameters and control parameters were used for all environments (see Appendix E for more details). This further demonstrates the generality and robustness of our method.

Since KeyCLD is trained directly on image observations, quantitative metrics can only be expressed in the image domain. The mean square error (MSE) in the image domain is not a good metric of long term prediction accuracy [12,17]. A model that trivially learns to predict a static image, which is the average of the dataset, learns no dynamics at all yet this model could report a lower MSE than a model that did learn the dynamics but started drifting from the groundtruth after some time. Therefore, we use the valid prediction time (VPT) score [11,33] which measures how long the predicted images stay close to the groundtruth images of a sequence:

$$\text{VPT} = \operatorname{argmin}_i[\text{MSE}(\mathbf{z}'^i, \mathbf{z}^i) > \epsilon] \quad (22)$$

where $\mathbf{z}^i$ are the groundtruth images, $\mathbf{z}'^i$ are the predicted images and $\epsilon$ is the error threshold. $\epsilon$ is determined separately for the different environments because it depends on the relative size in pixels of moving parts. We define it as the MSE of the averaged image of the respective validation dataset. Thus it is the lower bound for a model that would simply predict a static image. We present evaluations with the following ablations and baselines:

**KeyCLD**  The full framework as described in Sections 2 and 3.

**KeyLD**  The constraint function is omitted.

**KeyODE2**  A second order neural ODE modelling the acceleration is used instead of the Lagrangian prior. The keypoint estimator and renderer model are identical to KeyCLD.

**Lag-caVAE**  The model presented by Zhong and Leonard [12]. We adapted the model to the higher resolution and colour images.

**Lag-VAE**  The model presented by Zhong and Leonard [12]. We adapted the model to the higher resolution and colour images.

**HGN**  Hamiltonian Generative Network presented by Toth et al. [8].

**Table 2**
Valid prediction time (higher is better, equation (22)) in number of predicted frames (mean ± std) for the different models evaluated on the 50 sequences in the validation set. Lag-caVAE and Lag-VAE are only reported on the pendulum environment, since they are unable to model more than one moving body without segmented images. HGN is only reported on non-actuated systems, since it is incapable of modelling external forces and torques. KeyCLD achieves the best results on all benchmarks.

|          | # actuators |          | KeyCLD        | KeyLD        | KeyODE2      | Lag-caVAE   | Lag-VAE      | HGN        |
|----------|-------------|----------|---------------|--------------|--------------|-------------|--------------|------------|
| Pendulum | 0           | (Fig. 5) | **43.1 ± 9.7**  | 16.4 ± 11.3  | 19.1 ± 6.2   | 0.0 ± 0.0   | 10.8 ± 13.8  | 0.2 ± 1.4  |
|          | 1           | (Fig. 6) | **39.3 ± 9.8**  | 14.9 ± 7.9   | 12.0 ± 4.1   | 0.0 ± 0.1   | 8.0 ± 10.2   | –          |
| Cartpole | 0           | (Fig. 7) | **39.9 ± 7.4**  | 29.8 ± 11.2  | 29.5 ± 9.5   | –           | –            | 0.0 ± 0.0  |
|          | 1           | (Fig. 8) | **38.4 ± 8.7**  | 28.0 ± 9.7   | 24.4 ± 7.9   | –           | –            | –          |
|          | 2           | (Fig. 9) | **30.2 ± 10.7** | 23.9 ± 9.6   | 17.7 ± 8.2   | –           | –            | –          |
| Acrobot  | 0           | (Fig. 10)| **47.0 ± 6.0**  | 40.0 ± 7.9   | 34.3 ± 9.5   | –           | –            | 2.2 ± 6.9  |
|          | 1           | (Fig. 11)| **46.8 ± 4.6**  | 29.5 ± 6.3   | 33.0 ± 7.4   | –           | –            | –          |
|          | 2           | (Fig. 12)| **47.0 ± 3.5**  | 39.1 ± 9.9   | 30.8 ± 9.3   | –           | –            | –          |

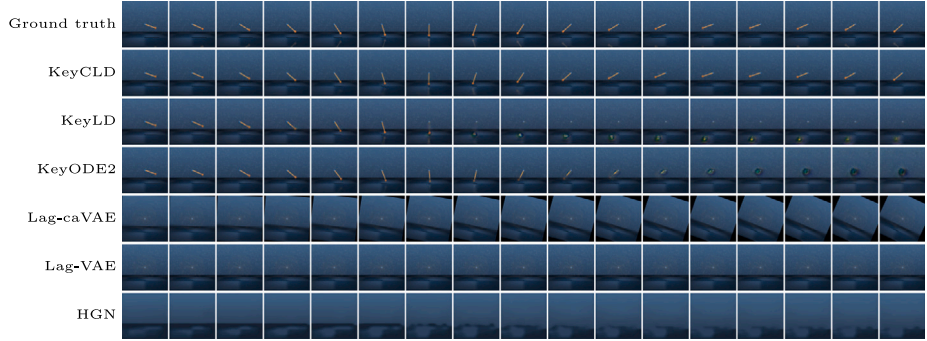

**Fig. 5.** Future frame predictions of the unactuated pendulum. These correspond to the first row in Table 2. 50 frames are predicted based on the first three frames of the ground truth sequence to estimate the velocity. Every third frame of every sequence is shown. KeyCLD is capable of making accurate long-term predictions with minimal drift of the dynamics. Without constraint function, KeyLD is not capable of making long-term predictions. Similarly, KeyODE2 is unable of making long-term predictions. Lag-caVAE is fundamentally incapable of modelling data with background information, since the reconstructed images are explicitly rotated. Lag-VAE does not succeed in modelling moving parts in the data, and simply learns to predict static images. HGN also does not capture the dynamics and only learns the background.

*4.1. Future frame predictions*

We generate predictions of 50 frames, given the first 3 frames of the ground truth sequences to estimate the initial velocity according to Eq. (18). The VPT metric is calculated for the 50 sequences in the validation set (see Appendix D for details) and averaged. See Table 2 for an overview of results. Lag-caVAE is unable to model data with background (see also Fig. 5). Despite our best efforts for implementation and training, Lag-VAE and HGN perform very poorly. The models are not capable of handling the relatively more challenging visual structure of dm_control environments. Removing the constraint function (KeyLD) has a detrimental effect on the ability to make long-term predictions. Results are comparable to removing the Lagrangian prior altogether (KeyODE2). This suggests that modelling dynamics in Cartesian coordinates coupled with keypoint representations is in itself a very strong prior and that the effect of the Lagrangian prior without constraints is minimal. This is consistent with recent findings by Gruver et al. [34]. However, using a Lagrangian formulation allows leveraging a constraint function, since a general neural ode model cannot make use of a constraint function. Thus, if a constraint function is available, the Lagrangian prior becomes much more powerful.

One sequence of each experiment is visualized in the paper. Please compare these qualitative results for the unactuated and actuated pendulum environment (Fig. 5, 6), unactuated, underactuated and fully actuated cartpole environment (Fig. 7, 8, 9) and unactuated, underactuated and fully actuated acrobot environment (Fig. 10, 11, 12). Every third frame of the sequence is shown.

*4.2. Learned potential energy models*

Since the potential energy $V$ is explicitly modelled, we can plot values throughout sequences of the state space. A sequence of images is processed by the learned keypoint estimator model, and the states are then used to calculate the potential energy with the learned potential energy model. Absolute values of the potential energy are irrelevant, since the potential is relative, but we gain insights by moving parts of the system separately. See Fig. 13 for results for the pendulum, Figs. 14 and 15 for the cartpole and Figs. 16 and 17 for the acrobot.

*4.3. Learned input matrix models*

Learning the input matrix $\mathbf{g}(\mathbf{x})$ is crucial for learning dynamics models with external inputs $\mathbf{u}$. We can visualize the vector basis that is represented by the input matrix, by drawing the vectors originating on their respective keypoints. See Fig. 18, 19 and 20 for the input matrices that are learned with our model. Each input corresponds to a vector base, visualized in different colours. The vectors multiplied by their respective input, can be interpreted as forces acting on the keypoints. These qualitative results allow further insight in our method.

*4.4. Energy shaping control*

A major argument in favour of expressing dynamics in terms of a mass matrix and potential energy is the straightforward control design via passivity based control and energy shaping [35]. Our framework allows design of a simple energy shaping controller, see Appendix C for details and derivation. Fig. 21 shows results of successful swing-up of the pendulum, cartpole and acrobot system. The same control parameters $k_p = 5.0$ and $k_d = 2.0$ are used for all systems, demonstrating the generality of the control method.

## 5. Conclusion and future work

We introduce the use of keypoints to learn Lagrangian dynamics from images. Learned keypoint representations derived from images are directly used as positional state vector for learning constrained

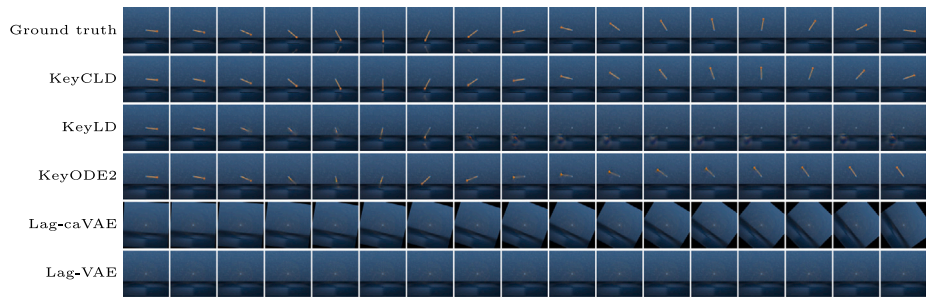

**Fig. 6.** Actuated pendulum.

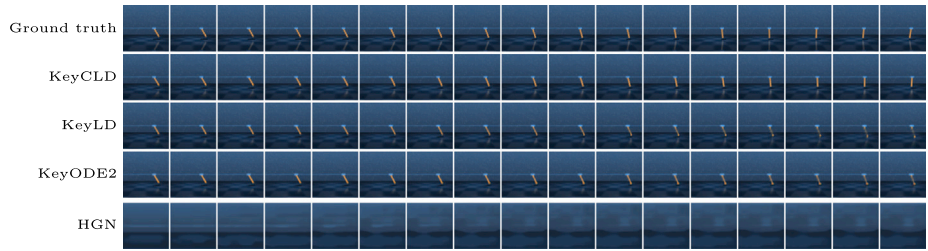

**Fig. 7.** Unactuated cartpole.

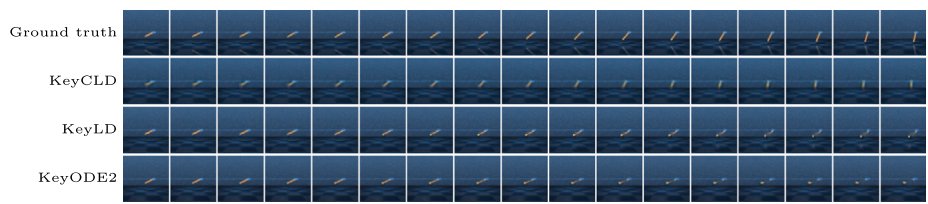

**Fig. 8.** Underactuated cartpole.

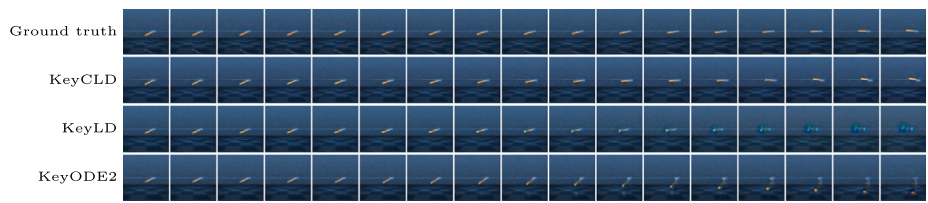

**Fig. 9.** Fully actuated cartpole.

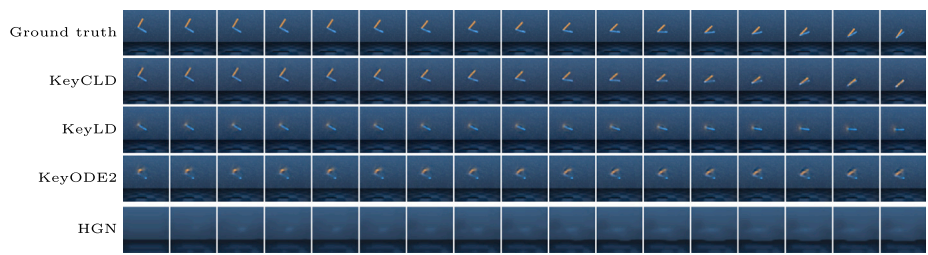

**Fig. 10.** Unactuated acrobot.

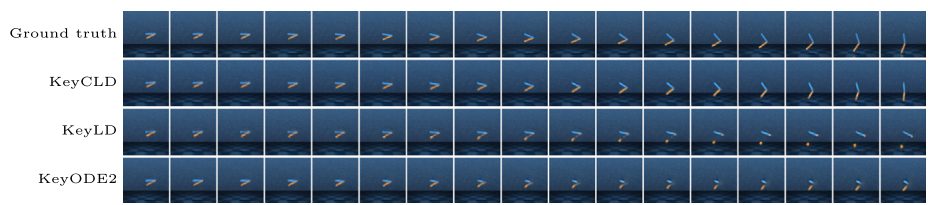

**Fig. 11.** Underactuated acrobot.

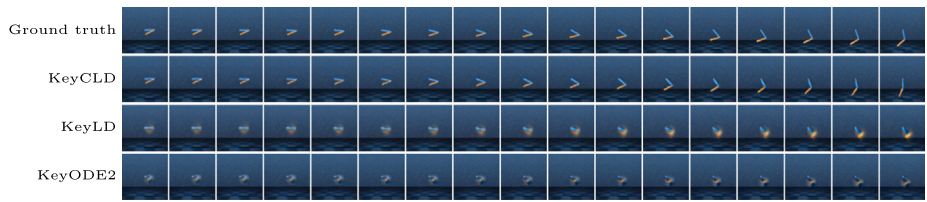

**Fig. 12.** Fully actuated acrobot.

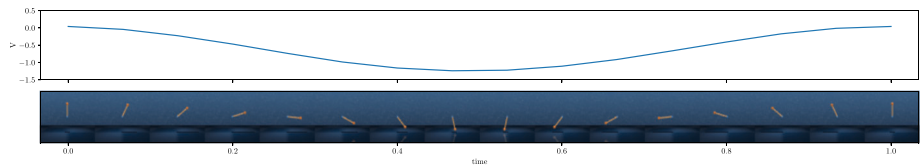

**Fig. 13.** Potential energy of the trained KeyCLD model of the pendulum environment. The pendulum makes a full rotation. As expected, the potential energy follows a smooth sinusoidal path throughout this sequence. The maximum value is reached when the pendulum is upright, and the minimum value is reached when the pendulum is down.

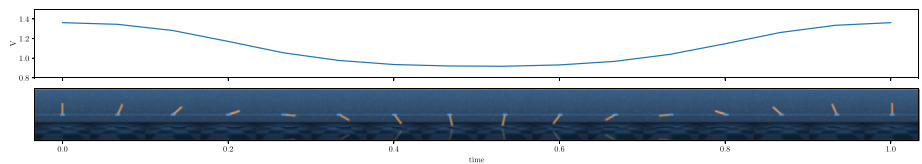

**Fig. 14.** Potential energy of the trained KeyCLD model of the cartpole environment. The position of the cart is fixed, and the pole makes a full rotation. As expected, the potential energy follows a smooth sinusoidal path throughout this sequence.

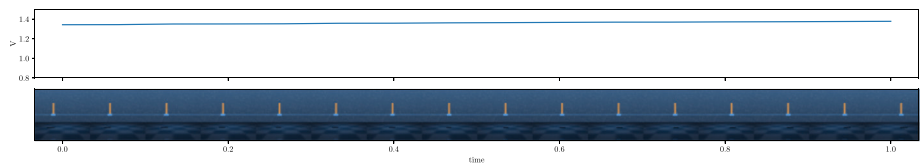

**Fig. 15.** Potential energy of the trained KeyCLD model of the cartpole environment. The pole is fixed and the cart moves from left to right. As expected, the change in potential energy in this sequence is very low (compare to Fig. 14 with the same axis). A horizontal movement has no impact on the gravity potential.

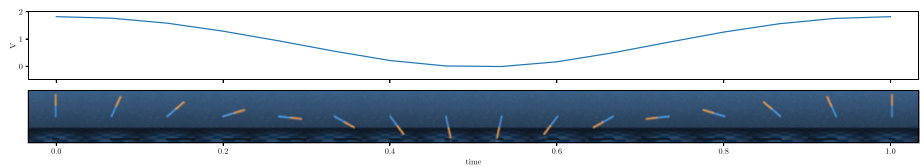

**Fig. 16.** Potential energy of the trained KeyCLD model of the acrobot environment. The first link makes a full rotation, the second link is fixed relative to the first link. As expected, the potential energy follows a smooth sinusoidal path throughout this sequence.

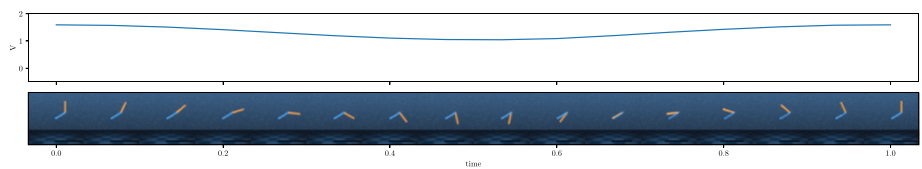

**Fig. 17.** Potential energy of the trained KeyCLD model of the acrobot environment. The first link is fixed and the second link makes a full rotation. Again, the potential energy follows a smooth sinusoidal path throughout this sequence. Please compare with Fig. 16, where both links are moving. Here the potential energy changes less, because the first link is not moving.

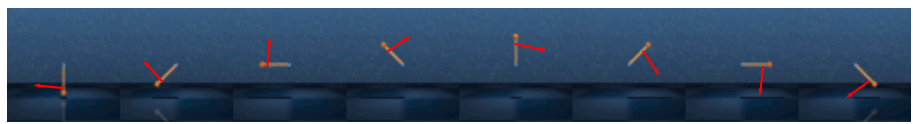

**Fig. 18.** Visualization of the input matrix of the trained KeyCLD model of the pendulum environment. The input of this environment is a torque acting on the pendulum. In the KeyCLD framework this is correctly modelled with a force acting perpendicular on the pendulum.

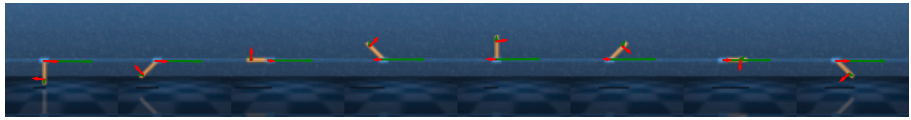

**Fig. 19.** Visualization of the input matrix of the trained KeyCLD model of the cartpole environment. This environment has two inputs, a horizontal force acting on the cart, and a torque acting on the pole. The horizontal force corresponds to the green vectors. The first vector acting on the cart keypoint stays constant, and the second vector is negligibly small, since the horizontal force does not act on the pole. The torque corresponds to the red vectors, it is modelled with forces acting on the pole in opposite directions, such that the residual force can be zero.

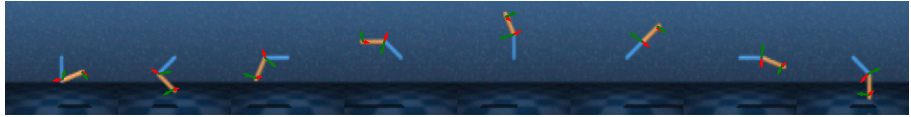

**Fig. 20.** Visualization of the input matrix of the trained KeyCLD model of the acrobot environment. This environment has two inputs, two torques acting on each pole. The torques are modelled with opposite forces on each end of the poles, such that the residual force can be zero.

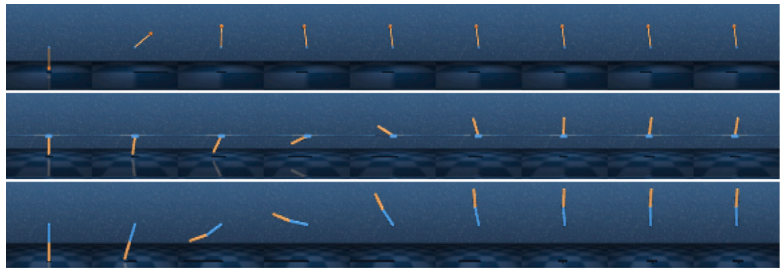

**Fig. 21.** KeyCLD allows using energy shaping control because the learned potential energy model is available. Based on a swing-up target image $\mathbf{z}^*$, the target state $\mathbf{x}^*$ is determined by the keypoint detector model. The sequences show that all three systems can achieve the target state. The control parameters $k_p = 5.0$ and $k_d = 2.0$ are the same for all systems, demonstrating the generality of the control method.

Lagrangian dynamics. The pendulum, cartpole and acrobot systems of dm_control are adapted as benchmarks. Previous works in literature on learning Lagrangian or Hamiltonian dynamics from images were benchmarked on very simple renderings of flat sprites on blank backgrounds, whereas dm_control is rendered with lighting effects, shadows, reflections and backgrounds. Also the recently proposed benchmarks by Botev et al. [11] have minimalistic visuals and do not cover a control aspect or external forces. We believe that working towards more realistic datasets is crucial for applying Lagrangian models in the real world.

The challenge of learning Lagrangian dynamics from more complex images should not be underestimated. Despite our best efforts in implementing and training Lag-caVAE, Lag-VAE [12] and HGN [8], they perform very poorly on our dm_control quantitative benchmark (see Table 2). KeyCLD is capable of making long-term predictions and learning accurate energy models, suitable for simple energy shaping control. KeyCLD is compared to a Lagrangian dynamics model without constraints (KeyLD) and a general second order neural ODE (KeyODE2). Both yield worse results on the benchmark, with comparable results (see Table 2). This suggests that when no constraint prior is available, the benefit of a Lagrangian prior is limited. When the constraint function is known, the constrained Lagrangian formulation is very beneficial for long-term predictions.

The main limitations of our work are that we only consider 2D systems, where the plane of the system is parallel with the camera plane. Secondly, we do not model energy dissipation, all systems have perfect energy conservation. Finally, the constraint function is given and we benchmark on relatively simple dynamical systems with few degrees of freedom.

Elevation to 3D, e.g. setups with multiple cameras, is an interesting future direction. Modelling contacts by using inequality constraints could also be a useful addition. And modelling energy dissipation is necessary for real-world applications. Several recent papers have proposed methods to incorporate energy dissipation in the Lagrangian dynamics models [6,36]. However, Gruver et al. [34] argue that modelling the acceleration directly with a second order differential equation and expressing the system in Cartesian coordinates, is a better approach. Further research into both approaches would clarify the benefit of Lagrangian and Hamiltonian priors on real-world applications. Applications on more complex scenarios with more degrees of freedom are also interesting for future work.

## 6. Broader impact

A tenacious divide exists between control engineering researchers and computer science researchers working on control. Where the first would use known equations of motion for a specific class of systems and investigate system identification, the latter would strive for the most general method with no prior knowledge. We believe this is a spectrum worth exploring, and as such use strong physics priors as Lagrangian mechanics, but still model e.g. the input matrix and the potential energy with arbitrary neural networks. The broad field of model-based reinforcement learning could benefit from decades of theory and practice in classic control theory and system identification. We hope this paper could help bridge both worlds.

Using images as input is, in a broad sense, very powerful. Since camera sensors are consistently becoming cheaper and more powerful due to advancements in technology and scaling opportunities, we can leverage these rich information sources for a deeper understanding of the world our intelligent agents are acting in. Image sensors can replace and enhance multiple other sensor modalities, at a lower cost.

To conclude, this work demonstrates the ability to efficiently model and control dynamical systems that are captured by cameras, with no supervision and minimal prior knowledge. We want to stress that we have shown it is possible to learn both the Lagrangian dynamics and state estimator model from images in one end-to-end process. The complex interplay between both, often makes them the most labour intensive parts in system identification. We believe this is a gateway step in achieving reliable end-to-end learned control from pixels.

**CRediT authorship contribution statement**

**Rembert Daems:** Conceptualization, Methodology, Software, Writing – original draft. **Jeroen Taets:** Conceptualization, Methodology, Writing – original draft. **Francis wyffels:** Conceptualization, Writing – review & editing, Supervision, Project administration, Funding acquisition. **Guillaume Crevecoeur:** Conceptualization, Writing – review & editing, Supervision, Project administration, Funding acquisition.

**Declaration of competing interest**

The authors declare the following financial interests/personal relationships which may be considered as potential competing interests: This research received funding from the Flemish Government under the "Onderzoeksprogramma Artificiële Intelligentie (AI) Vlaanderen" programme. Furthermore it was supported by Flanders Make under the SBO projects MultiSysLeCo and CADAIVISION.

**Data availability**

Data will be made available on request.

**Acknowledgments**

The authors thank Jonas Degrave, Peter De Roovere and Tom Lefebvre for many insightful discussions.

**Appendix A. Implementation of constrained Euler–Lagrange equations in JAX**

It could seem a daunting task to implement the derivation of the constrained Euler–Lagrange Eqs. (13) in an autograd library. As an example, we provide an implementation in JAX [23].

```python
import jax
import jax.numpy as jnp

def constraint_fn(x):
    # function that returns a vector with constraint values
    c = jnp.array([
        ...,
    ])
    return c

def mass_matrix(params, x):
    # function that returns the mass matrix
    ...
    return m

def potential_energy(params, x):
    # function that returns the potential energy
    ...
    return V

def input_matrix(params, x):
    # function that returns the input matrix
    ...
    return g

def euler_lagrange(params, x, x_t, action):
    m_inv = jnp.linalg.pinv(mass_matrix(params, x))
    f = - jax.grad(potential_energy, 1)(params, x) + \
        input_matrix(params, x) @ action

    Dphi = jax.jacobian(constraint_fn)(x)
    DDphi = jax.jacobian(jax.jacobian(constraint_fn))(x)

    # Lagrange multiplicators:
    l = jnp.linalg.pinv(Dphi @ m_inv @ Dphi.T) @ (Dphi @ m_inv @ f +
        DDphi @ x_t @ x_t)
    x_tt = m_inv @ (f - Dphi.T @ l)

    return x_tt
```

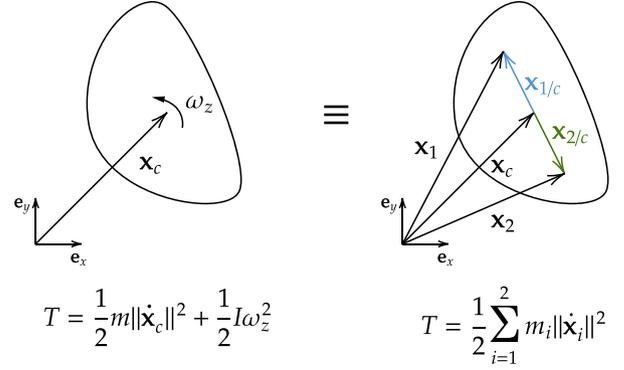

**Fig. B.22.** Any 2D rigid body with mass $m$ and rotational inertia $I$ is equivalent to a set of two point masses $\mathbf{x}_1$ and $\mathbf{x}_2$ with masses $m_1$ and $m_2$. The kinetic energy of the rigid body, expressed in a translational part and a rotational part, is equal to the sum of the kinetic energies of the point masses.

**Appendix B. Rigid bodies as sets of point masses**

The position of a rigid body in 2D is fully described by the position of its center of mass $\mathbf{x}_c$ and orientation $\theta$. Potential energy only depends on the position, thus if we want to describe the potential energy with an equivalent rigid set of point masses, two points are sufficient to fully determine $\mathbf{x}_c$ and $\theta$. For the kinetic energy, we provide the following Theorem and proof:

**Theorem 1.** *For any 2D rigid body, described by its center of mass $\mathbf{c}$, mass $m$ and rotational inertia $I$, there exists an equivalent rigid set of two point masses $\mathbf{x}_1$ and $\mathbf{x}_2$ with masses $m_1$ and $m_2$.*

**Proof.** To find conditions such that the kinetic energy expressed in two point masses should be equal to the rigid body representation, we start by expressing general 3D-movement:

$$\mathbf{x}_i = \mathbf{x}_c + \mathbf{x}_{i/c} \qquad , i \in \{1, 2\} \tag{B.1}$$

Where the vector $\mathbf{x}_c$ are the coordinates of the center of mass and the vector $\mathbf{x}_{i/c}$ is the position of the point mass relative to the center of mass. Since this relative position $\mathbf{x}_{i/c}$ has fixed length, only a rotation is possible and hence the equation of the velocity is:

$$\dot{\mathbf{x}}_i = \dot{\mathbf{x}}_c + \boldsymbol{\omega} \times \mathbf{x}_{i/c} \qquad , i \in \{1, 2\} \tag{B.2}$$

where $\boldsymbol{\omega}$ is the rotational velocity of the body. Substituting this in the kinetic energy of the point masses, we get:

$$\begin{aligned}
T &= \frac{1}{2} \sum_{i=1}^{2} m_i \| \dot{\mathbf{x}}_c + \boldsymbol{\omega} \times \mathbf{x}_{i/c} \|^2 \\
&= \frac{1}{2} \sum_{i=1}^{2} m_i \left( \| \dot{\mathbf{x}}_c \|^2 + \| \boldsymbol{\omega} \times \mathbf{x}_{i/c} \|^2 + 2 \mathbf{x}_{i/c} \cdot (\dot{\mathbf{x}}_c \times \boldsymbol{\omega}) \right)
\end{aligned} \tag{B.3}$$

Where we calculated the square and used the circular shift property of the triple product on the last term.

For movement in the 2D-plane (i.e. $\boldsymbol{\omega} = \vec{\mathbf{e}}_z \omega_z$ and $\mathbf{x}_i = \vec{\mathbf{e}}_x x_{i,x} + \vec{\mathbf{e}}_y x_{i,y}$), this becomes:

$$\begin{aligned}
T &= \frac{1}{2} \sum_{i=1}^{2} m_i \left( \| \dot{\mathbf{x}}_c \|^2 + \| \mathbf{x}_{i/c} \|^2 \omega_z^2 + 2 \mathbf{x}_{i/c} \cdot (\dot{\mathbf{x}}_c \times \boldsymbol{\omega}) \right) \\
&= \frac{1}{2} (m_1 + m_2) \| \dot{\mathbf{x}}_c \|^2 + \frac{1}{2} \left( m_1 \| \mathbf{x}_{1/c} \|^2 + m_2 \| \mathbf{x}_{2/c} \|^2 \right) \omega_z^2 + \big( m_1 \mathbf{x}_{1/c} \\
&\quad + m_2 \mathbf{x}_{2/c} \big) \cdot (\dot{\mathbf{x}}_c \times \boldsymbol{\omega})
\end{aligned} \tag{B.4}$$

Matching the kinetic energy of the 2 point masses (Eq. (B.4)) with that of the rigid body representation (left hand side of Fig. B.22), we get following conditions:

$$\begin{cases} m = m_1 + m_2 \\ I = m_1 \|\mathbf{x}_{1/c}\|^2 + m_2 \|\mathbf{x}_{2/c}\|^2 \\ \mathbf{0} = m_1 \mathbf{x}_{1/c} + m_2 \mathbf{x}_{2/c} \end{cases} \quad (B.5)$$

Since the last equation is a vector equation, this gives us four equations in six unknowns ($m_1$, $m_2$, $\mathbf{x}_{1,x}$, $\mathbf{x}_{1,y}$, $\mathbf{x}_{2,x}$, $\mathbf{x}_{2,y}$), which leaves us the freedom to choose two. □

It follows from the third condition of (B.5) that points $\mathbf{x}_1$, $\mathbf{x}_2$ and $\mathbf{x}_c$ should be collinear. To conclude, we can freely choose the positions of the point masses (as long as $\mathbf{x}_c$ is on the line between them), and will be able to model the rigid body as a set of two point masses. In practice, KeyCLD will freely choose the keypoint positions to be able to model the dynamics. Depending on the constraints in the system, it is possible to further reduce the number of necessary keypoints. See Appendix D for examples.

## Appendix C. Energy shaping control

Recent works of Zhong et al. [6,12] use energy shaping in generalized coordinates. In Cartesian coordinates, energy shaping can still be used. This is easily seen from the fact that for the holonomic constraints $\Phi(\mathbf{x}) \equiv \mathbf{0}$, we have the derivative $D\Phi(\mathbf{x})\dot{\mathbf{x}} = \mathbf{0}$, which means that the constraint forces in Eq. (13) are perpendicular to the path and hence do no work nor influence the energy [22].

Energy shaping control makes sure that the controlled system behaves according to a potential energy $V_d(\mathbf{x})$ instead of $V(\mathbf{x})$:

$$\mathbf{u} = (\mathbf{g}^\top \mathbf{g})^{-1} \mathbf{g}^\top (\nabla_{\mathbf{x}} V - \nabla_{\mathbf{x}} V_d) - y_{\text{passive}} \quad (C.1)$$

where $y_{\text{passive}}$ can be any passive output, the easiest choice being $y_{\text{passive}} = k_d \mathbf{g}^\top \dot{\mathbf{x}}$, where $k_d$ is a tuneable control parameter. The proposed potential energy $V_d$ should be such that:

$$\mathbf{x}^* = \arg\min V_d(\mathbf{x})$$
$$\mathbf{0} = \mathbf{g}^\perp (\nabla_{\mathbf{x}} V - \nabla_{\mathbf{x}} V_d) \quad (C.2)$$

Where $\mathbf{g}^\perp$ is the left-annihilator of $\mathbf{g}$, meaning that $\mathbf{g}^\perp \mathbf{g} = \mathbf{0}$. For fully actuated systems, the first condition of Eq. (C.2) is always met and the easiest choice is:

$$V_d(\mathbf{x}) = (\mathbf{x} - \mathbf{x}^*)^\top k_p (\mathbf{x} - \mathbf{x}^*) \quad (C.3)$$

where $k_p$ is a tuneable control parameter. The desired equilibrium position $\mathbf{x}^*$ is obtained by processing an image of the desired position with the keypoint estimator model. Finally, the passivity-based controller that is used is:

$$\mathbf{u} = (\mathbf{g}^\top \mathbf{g})^{-1} \mathbf{g}^\top \left[ \nabla_{\mathbf{x}} V - k_p (\mathbf{x} - \mathbf{x}^*) \right] - k_d \mathbf{g}^\top \dot{\mathbf{x}} \quad (C.4)$$

Changing the behaviour of the kinetic energy is also possible [37], but is left for future work. Many model-based reinforcement learning algorithms require the learning of a full neural network as controller. Whilst in this work, due to knowledge of the potential energy, we only need to tune two parameters $k_p$ and $k_d$.

## Appendix D. Details about the `dm_control` environments and data generation

We adapted the pendulum, cartpole and acrobot environments from `dm_control` [20] implemented in MuJoCo [32]. Both are released under the Apache-2.0 license. Following changes were made to the environments to adapt them to our use-case:

*Pendulum.* The camera was repositioned so that it is in a parallel plane to the system. Friction was removed. Torque limits of the motor are increased.

*Cartpole.* The camera was moved further away from the system to enable a wider view, the two rails are made longer and the floor lowered so that they are not cut-off with the wider view. All friction is removed. The pole is made twice as thick, the colour of the cart is changed. Torque limits are increased and actuation is added to the cart to make full actuation possible.

*Acrobot.* The camera and system are moved a little bit upwards. The two poles are made twice as thick, and one is changed in colour. Torque limits are increased and actuation is added to the upper part to make full actuation possible.

*Data generation.* For every environment, 500 runs of 50 timesteps are generated with a 10% validation split. The initial state for every sequence is at a random position with small random velocity. The control inputs **u** are constant throughout a sequence, and uniform randomly chosen between the force and torque limits of the input. We set $\mathbf{u} = \mathbf{0}$ for 20% of the sequences. We found this helps the model to learn the dynamics better, discouraging confusion of the energy models with external actions.

The constraint function for each of the environments are given in Fig. D.23. As explained in Appendix B, every rigid body needs to be represented by two keypoints. But due to the constraints it is possible to omit certain keypoints, because they do not move or coincide with other keypoints. As experimentally validated, we can thus model all three systems with a lower number of keypoints, where the number of keypoints equals the number of bodies.

*Pendulum.* One keypoint is used to model the pendulum. The second keypoint of this rigid body can be omitted because it can be assumed to be at the origin. Due to the constraint function, this point will provide no kinetic energy since it will not move. Since the other keypoints position and mass is freely chosen, any pendulum can be modelled. The constraint function expresses that the distance $l_1$ from the origin to $\mathbf{x}_1$ is fixed. The value of $l_1$ in the implementation is irrelevant because it vanishes when taking the Jacobian.

*Cartpole.* Two keypoints are used to model the cartpole. The constraint function expresses that $\mathbf{x}_1$ does not move in the vertical direction and the distance $l_1$ between $\mathbf{x}_1$ and $\mathbf{x}_2$ is constant. Again, the values of $l_1$ and $l_2$ in the implementation are irrelevant.

*Acrobot.* Two keypoints are used to model the acrobot. The constraint function expresses that lengths $l_1$ and $l_2$ are constant through time. Again, the values are irrelevant in the implementation.

## Appendix E. Training hyperparameters and details

All models were trained on one NVIDIA RTX 2080 Ti GPU.

*KeyCLD, KeyLD and KeyODE2.* We use the Adam optimizer [38], implemented in Optax [39] with a learning rate of $3 \times 10^{-4}$. We use the exact same hyperparameters for all the environments and did not tune them individually. Dynamics loss weight $\lambda = 1$, $\sigma = 0.1$ for the Gaussian blobs in $\mathbf{s}'$. The hidden layers in the keypoint estimator and renderer model have at the first block 32 features, this increases to respectively 64 and 128 after every maxpool operation. All convolutions have kernel size $3 \times 3$, and maxpool operations scale down with factor 2 with a kernel size of $2 \times 2$. See Tables E.5 and E.6 for the number of parameters.

The potential energy is modelled with an MLP with two hidden layers with 32 neurons and celu activation functions [40]. The weights are initialized with a normal distribution with standard deviation 0.01. See Table E.3 for the number of parameters. Likewise, the input matrix is modelled with an MLP similar to the potential energy. The outputs of this MLP are reshaped in the shape of the input matrix. See Table E.4 for the number of parameters.

The KeyODE2 dynamics model is an MLP with three hidden layers with each 64 neurons. We chose a higher number of layers and neurons, to allow this model more expressivity compared to the potential energy and input matrix models of KeyCLD.

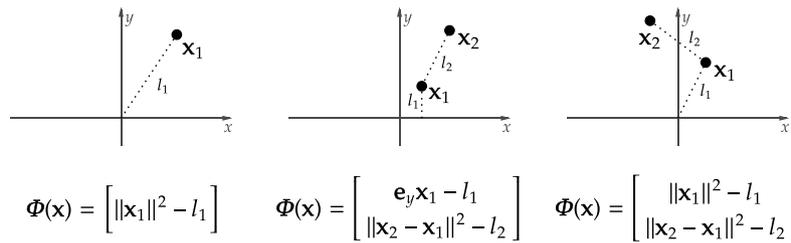

**Fig. D.23.** From left to right the pendulum, cartpole and acrobot `dm_control` environments. The respective constraint functions are given below each schematic.

**Table E.3**
Number of parameters of the potential energy model.

|         | Pendulum | Cartpole | Acrobot |
|---------|---------:|---------:|--------:|
| `Dense_0` | 96   | 160  | 160  |
| `Dense_1` | 1056 | 1056 | 1056 |
| `Dense_2` | 33   | 33   | 33   |
| Total   | 1185 | 1249 | 1249 |

**Table E.4**
Number of parameters of the input matrix model.

|         | Pendulum | Cartpole | Acrobot |
|---------|---------:|---------:|--------:|
| `Dense_0` | 96   | 160  | 160  |
| `Dense_1` | 1056 | 1056 | 1056 |
| `Dense_2` | 66   | 264  | 264  |
| Total   | 1218 | 1480 | 1480 |

**Table E.5**
Number of parameters of the keypoint encoder model.

|         | Pendulum | Cartpole | Acrobot |
|---------|---------:|---------:|--------:|
| `Block_0/Conv_0`      | 896    | 896    | 896    |
| `Block_0/GroupNorm_0` | 64     | 64     | 64     |
| `Block_1/Conv_0`      | 18 496 | 18 496 | 18 496 |
| `Block_1/GroupNorm_0` | 128    | 128    | 128    |
| `Block_2/Conv_0`      | 73 856 | 73 856 | 73 856 |
| `Block_2/GroupNorm_0` | 256    | 256    | 256    |
| `Block_3/Conv_0`      | 110 656| 110 656| 110 656|
| `Block_3/GroupNorm_0` | 128    | 128    | 128    |
| `Block_4/Conv_0`      | 27 680 | 27 680 | 27 680 |
| `Block_4/GroupNorm_0` | 64     | 64     | 64     |
| `Conv_0`              | 289    | 578    | 578    |
| Total                 | 232 513| 232 802| 232 802|

*Lag-caVAE, Lag-VAE and HGN.* For the Lag-caVAE and Lag-VAE baselines, the official public codebase was used [12]. We adapted the implementation to work with the higher input resolution of 64 by 64 (instead of 32 by 32), and 3 colour channels (instead of 1).

For the HGN baseline, we used the implementation that was also released by Zhong and Leonard [12]. The architecture was adapted to work with the higher input resolution of 64 by 64 (instead of 32 by 32) by adding an extra upscale layer in the decoder, and a maxpool layer and one extra convolutional layer in the encoder.

**Table E.6**
Number of parameters of the renderer model.

|         | Pendulum | Cartpole | Acrobot |
|---------|---------:|---------:|--------:|
| `Seed Tensor`         | 126 976 | 126 976 | 122 880 |
| `Block_0/Conv_0`      | 9248    | 9248    | 9248    |
| `Block_0/GroupNorm_0` | 64      | 64      | 64      |
| `Block_1/Conv_0`      | 18 496  | 18 496  | 18 496  |
| `Block_1/GroupNorm_0` | 128     | 128     | 128     |
| `Block_2/Conv_0`      | 73 856  | 73 856  | 73 856  |
| `Block_2/GroupNorm_0` | 256     | 256     | 256     |
| `Block_3/Conv_0`      | 110 656 | 110 656 | 110 656 |
| `Block_3/GroupNorm_0` | 128     | 128     | 128     |
| `Block_4/Conv_0`      | 27 680  | 27 680  | 27 680  |
| `Block_4/GroupNorm_0` | 64      | 64      | 64      |
| `Conv_0`              | 867     | 867     | 867     |
| Total                 | 368 419 | 364 323 | 364 323 |

### Appendix F. Failure cases

A possible failure case is that the model learns a faulty keypoint representation that does not correspond to the given constraint function. This results in a failed model, and the training is stuck in this local minima. The encoder model will keep focussing on this erronous representation and is unable to switch to the correct keypoints. Fig. F.24 shows an example of this failure case. We observed this failure in a minority of the experiments. This can be mitigated by retraining the model.

### Appendix G. Ablating $L_e$

A binary cross-entropy loss $L_e$ is formulated over **s** and **s'** to encourage the Gaussian prior. When $L_e$ is omitted, the model can get stuck in a local minima where the encoder does not learn to predict keypoints, but rather larger regions or static values. Image G.25 shows an example of this failure case.

### Appendix H. Supplementary data

Supplementary material related to this article can be found online at https://doi.org/10.1016/j.neucom.2023.127175.

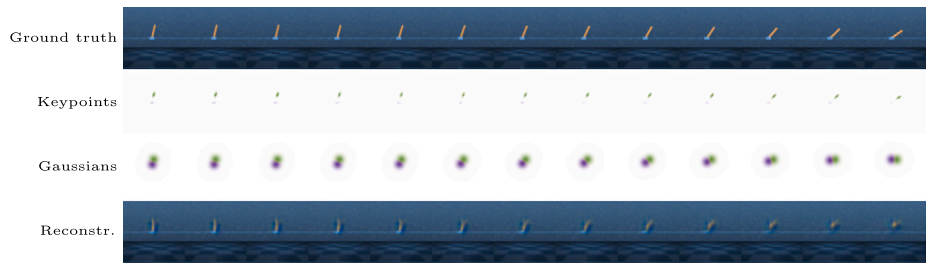

**Fig. F.24.** Failure case of KeyCLD on the cartpole environment. Keypoints are indicated in green and purple. The model erroneously assigned the green keypoint to the pole. Since the given constraint function dictates that the green keypoint can only move horizontally, this results in faulty model.

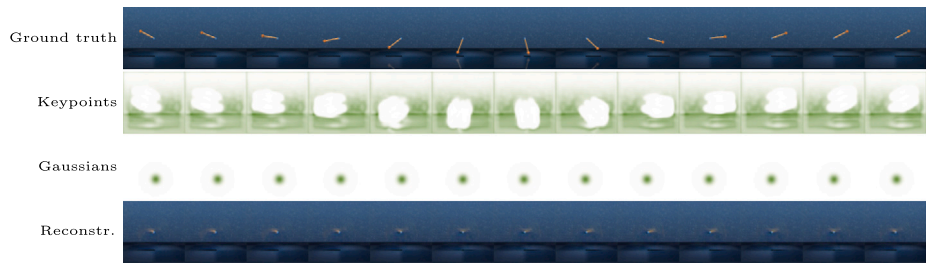

**Fig. G.25.** Omitting $L_e$ can result in poor learning of keypoints. The encoder model does not predict distinct keypoints, but other shapes. This is effectly a local minima in the learning process, since the model is uncapable of switching to a correct representation.

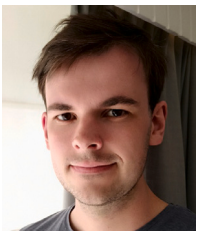

**Rembert Daems** (°1991) received his Master degree in electromechanical engineering from Ghent University (2016). He started working at CNH Industrial and subsequently at Octinion. In 2019 he fully returned to academia to work on a Ph.D. at Ghent University. He studies at D2LAB under the guidance of Guillaume Crevecoeur and at DILab - AIRO under the guidance of Francis wyffels. His interests are computer vision and dynamics.

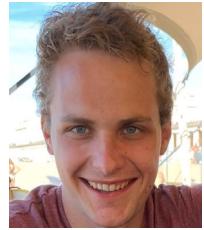

**Jeroen Taets** obtained an M.Sc. in mechanical-electrical engineering from Ghent University (2019). He then started a Ph.D. at UGent in D2LAB in the research group of Prof. Guillaume Crevecoeur. His research interests are data-driven control techniques for realizing high-performance mechatronic systems like energy-based Reinforcement Learning, hybrid optimal control, data-driven autotuning and more.

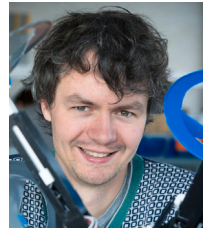

**Francis wyffels** obtained a Ph.D. in Computer Science in 2013. Since 2016, he has been a professor in machine learning and robotics at the AI & Robotics Lab (IDLab-AIRO) at Ghent University – imec. Over the last decade, he has built expertise in training recurrent neural networks, robotics, and unconventional computing with both living plants and robots. Together with his team of 16 researchers, he works towards a general-purpose robot that can perform everyday tasks in your home or workspace. In this research, the link between (robot) morphology, control and perception are central.

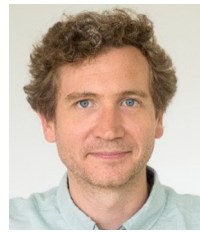

**Guillaume Crevecoeur** (°1981) received his Master and Ph.D. degree in Engineering Physics from Ghent University in 2004 and 2009, respectively. He received a Research Foundation Flanders postdoctoral fellowship in 2009 and was appointed Associate Professor at the Department of Electromechanical, System and Metal Engineering from the Ghent University in 2014. He currently leads the Ghent University activities on Machines, Intelligence, Robotics and electrOmechanical systems within the Flanders Make. With his team, he conducts research at the intersection of system identification, control and machine learning for mechatronic, industrial robotic, and energy systems. His goal is to endow physical dynamic systems with improved functionalities and capabilities when interacting with uncertain environments, other systems and humans.